\documentclass{article}
\usepackage[utf8]{inputenc}
\usepackage{authblk}
\usepackage{setspace}
\usepackage[margin=1.25in]{geometry}
\usepackage{graphicx}
\graphicspath{ {./figures/} }
\usepackage{subcaption}
\usepackage{times}
\usepackage{amsmath}
\usepackage{amssymb}
\usepackage{makecell}
\usepackage{authblk}
\usepackage{hyperref}

\usepackage[table, svgnames]{xcolor} % merge any needed options here
\usepackage[most]{tcolorbox}
\definecolor{lightgray}{HTML}{F6F6F6}
\definecolor{mediumgray}{HTML}{C6CACA}
\newtcolorbox{examplebox}[1][]{
  colback=lightgray,      
  colframe=mediumgray,    
  boxrule=0.8pt,
  arc=2pt,
  left=6pt,
  right=6pt,
  top=6pt,
  bottom=6pt,
  fonttitle=\color{black},
  title=#1                      
}

\usepackage[style=ieee, citestyle=numeric-comp, sorting=none]{biblatex}
\title{Structure Over Signal: A Globalized Approach to Multi-relational GNNs for Stock Prediction}

\author[1]{Amber Li}
\author[1]{Aruzhan Abil}
\author[1]{Juno Marques Oda}

\affil[1]{Department of Mathematics, Columbia University, New York, USA

{\small \texttt{\{al4580, aa5563, jam2528\}@columbia.edu}}}

\begin{document}
\maketitle

%%%%%% Abstract %%%%%%
\begin{abstract}
In financial markets, Graph Neural Networks have been successfully applied to modeling relational data, effectively capturing nonlinear inter-stock dependencies. Yet, existing models often fail to efficiently propagate messages during macroeconomic shocks. In this paper, we propose OmniGNN, an attention-based multi-relational dynamic GNN that integrates macroeconomic context via heterogeneous node and edge types for robust message passing. Central to OmniGNN is a sector node acting as a global intermediary, enabling rapid shock propagation across the graph without relying on long-range multi-hop diffusion. The model leverages Graph Attention Networks (GAT) to weigh neighbor contributions and employs Transformers to capture temporal dynamics across multiplex relations. Experiments show that OmniGNN outperforms existing stock prediction models on public datasets, particularly demonstrating strong robustness during the COVID-19 period. \footnote{The code and models of OmniGNN are made publicly available at https://github.com/amberhli/OmniGNN.}
\\

\end{abstract}

%%%%%% Main Text %%%%%%
\section{Introduction}
In recent years, Graph Neural Networks (GNNs) have been established as the standard for prediction tasks on graph datasets, with significant applications in areas such as molecular modeling, transportation networks, and recommendation systems \cite{14, 15, 8, 2}. In the financial domain, recent fusion models that pair Graph Convolutional Networks (GCNs) with temporal Recurrent Neural Networks (RNNs) like Long Short-Term Memory (LSTM) or Gated Recurrent Units (GRUs) have set the benchmark for dynamic node-level prediction tasks \cite{1, 5, 6}. These models, however, face several limitations. First, message-passing GCNs that rely on one-hop neighborhood aggregation fail to distinguish nonisomorphic graphs with similar local structures. Additionally, the standard GCN aggregation scheme applies uniform averaging to incoming messages, which fails to acknowledge the varying relevance of such messages \cite{3}. Second, although RNNs are well-suited for capturing long-range temporal dependencies, their sequential nature causes memory bottlenecks with the increase in complexity, limiting their ability to integrate information across the full temporal history \cite{9}. Third, many GNNs face the challenge of oversmoothing, which occurs when multiple rounds of message-passing leads to homogeneous node embeddings \cite{10}. While skip connections may be used to preserve node-level information during the update step, this strategy increases the parameter count while still leaving the graph susceptible to oversmoothing \cite{16}.

To address these challenges, we introduce \textbf{OmniGNN}, a novel GAT-Transformer fusion model that captures long-range dependencies and mitigates oversmoothing through three core contributions. 

First, we propose \textit{Metapath Attention-Weighting} for modeling relationship structures. Specifically, we define a set of metapaths that begin and end at stock nodes while traversing intermediate nodes of varying types (e.g., industries or regulatory entities). These metapaths capture higher-order semantic relationships that extend beyond stock-to-stock dependencies—for example,  structural correlations arising from shared industry affiliations. By learning the metapath attention weights during message aggregation, \hbox{OmniGNN} produces more expressive, relation-specific embeddings. 

Second, we propose \textit{Temporal Encoding via Transformer} to capture long-range dependencies as the network evolves over time. Unlike RNNs, Transformers eliminate sequential recurrence, allowing each time step to attend to all others in parallel \cite{12}. We apply Attention with Linear Biases (ALiBi) to the Transformer, introducing positional bias directly into attention score computation \cite{11}. This allows the model to prioritize recent observations in the graph network without learning positional embeddings, which leads to improved scalability. 

Third, we propose a novel \textit{``Global'' Node} for robust network topology. We design an industry node, or a global node connected to all stocks in that industry via multi-relational edges that reflect supply-chain, regulatory, and sectoral ties. This global node creates a star topology overlay such that any two nodes in the graph can communicate to each other in no more than 2 hops, shortening the message-passing path. Equipped with this virtual connection, our model can retain local information and capture topologically distant interactions without relying on as many GNN layers, effectively mitigating the oversmoothing challenge.

%%%%%%%%%%%%%%%%%%%%%%%%%%%%%%%%%%%%%%%%%%%%%%%%%%%%%%%%%%%%%%%%%
%%%%%%%%%%%% MODEL ARCHITECTURE HEADER %%%%%%%%%%%%%%%%%%%%%%%%%%
%%%%%%%%%%%%%%%%%%%%%%%%%%%%%%%%%%%%%%%%%%%%%%%%%%%%%%%%%%%%%%%%%
\section{Model Architecture}

OmniGNN consists of three key components: 1) Structural Layer — encodes nodes using weighted metapath representations. 2) Temporal Layer — learns dynamic node representations across time windows. 3) Prediction Layer — outputs the next day excess return for each stock node on trading day $t$. The model architecture is visualized in Figure 1.
\subsubsection*{Problem Formulation}
We formulate excess return prediction as a linear regression task. A sequence of discrete, multi-relational graphs indexed by time $t$, from trading day $1$ to $T$, is defined as: $\mathcal{G} = \{\mathcal{G}^{(t)}\}_{t=1}^T$. Each graph snapshot is represented by $\mathcal{G}^{(t)} = {\mathcal{V}^{(t)}, \mathcal{A}^{(t)}, \mathcal{E}^{(t)}}$, where $\mathcal{V}^{(t)}$ denotes the set of stock and industry nodes, $\mathcal{A}^{(t)}$ is the adjacency tensor capturing node relationships, and $\mathcal{E}^{(t)}$ contains the corresponding multi-dimensional edge attributes.

Concretely, we represent a stock as $\mathcal{S}$ and industry as $\mathcal{I}$, and define their corresponding edges $\mathcal{E}_{\mathcal{SS}}$ and $\mathcal{E}_{\mathcal{SI}}$ using inter-market relations between stocks and industries.

%Let  be a sequence of discrete, multi-relational graphs indexed by %time $t$, from trading day $1$ to $T$. Each g, where:
%\begin{itemize}
 %   \item is the fixed set of $N$ nodes at time $t$. Each node $v_i \in \mathcal{V}^{(t)}$ is associated with a feature vector $x_i^{(t)}\in\mathbb{R}^F$ and a scalar label $y_i^{(t)}\in\mathbb{R}$, where $F=16$ is the node feature dimension.
    %\item $\mathcal{A}^{(t)} \in \{0, 1\}^{|\mathcal{R}| \times N \times N}$ is a multi-relational binary adjacency tensor encoding edge existence under relation types $\mathcal{R}=\{\mathcal{SS}, \mathcal{SI}\}$.
   %\item $\mathcal{E}^{(t)} \in \mathbb{R}^{|\mathcal{R}| \times N \times N \times E}$ stores multiple edge attributes for each relation type, where $E=2$ is the edge feature dimension.
%\end{itemize}
\subsubsection*{Edge Building}
Companies within the same sector often share common business practices and risk exposures. We use Global Industry Classification Standard (GICS) codes, which define hierarchical levels of sector classification, to compute the sector similarity between two stocks as follows \cite{18}:
\begin{equation}
    \text{SectorSim}(v_i, v_j) = \frac{L(v_i, v_j)}{4},
\end{equation}
where $L(v_i, v_j) \in \{0,1,2,3,4\}$ denotes the depth of the lowest common ancestor (LCA) of $v_i$ and $v_j$ on the GICS classification tree. A value of $0$ means the two stocks share no common sectoral classification, while a value of $4$ means $v_i$ and $v_j$ occupy the same sub-industry.\\

Moreover, institutional shareholders often hold positions in several stocks at once, which can lead to increased information flow and correlated trading behavior between such stocks. We calculate the ratio of overlapping shareholders, weighted by the value of their position on a given day:
\begin{equation}
    \text{ShareholderSim}(v_i,v_j) = \frac{\sum_{k \in S_i \cap S_j} \text{min}(w_{i, k}, w_{i, k})}{\sum_{k \in S_i \cup S_j} \text{max}(w_{i, k}, w_{i, k})},
\end{equation}
Together, these features compose the two-dimensional $\mathcal{SS}$ edge between two stocks. 
\begin{examplebox}[Stock-Stock Edge Definition]
$$\mathcal{E}_{\mathcal{SS}}^{(t)}(v_i, v_j)[0] = \text{SectorSim}(v_i, v_j),\qquad \mathcal{E}_{\mathcal{SS}}^{(t)}(v_i, v_j)[1] = \text{ShareholderSim}(v_i,v_j)$$
\end{examplebox}
A company's performance is closely tied to its industry, where competition and government policies can significantly affect stock outcomes. To capture these relationships, we calculate each stock's market share in a given sector using quarterly revenue data from Capital IQ\footnote{Source: S\&P Capital IQ} :
\begin{equation}
\text{MarketShare}(v_i) = \frac{\text{Revenue}(v_i)}{\sum_{j \in \mathcal{I}}\text{Revenue}(v_j)},
\end{equation}
and obtain each stock's monthly sector-specific Governance score—reflecting its Corporate Social Responsibility strategy—using ESG data from the London Stock Exchange.
These features compose the two-dimensional $\mathcal{SI}$ edge between a stock and its industry. 
\begin{examplebox}[Stock-Industry Edge Definition]
$$\mathcal{E}_{\mathcal{SI}}^{(t)}(v_i, v_j)[0] = \text{MarketShare}(v_i, v_j),\qquad \mathcal{E}_{\mathcal{SI}}^{(t)}(v_i, v_j)[1] = \text{ESGGovernance}(v_i,v_j)$$
\end{examplebox}

%%%%%%%%%%%%%%%%%%%%%%%%%%%%%%%%%%%%%%%%%%%%%%%%%%%%%%%%%%%%%%%%%
%%%%%%%%%%%% LAYER 1: STRUCTURAL LAYER %%%%%%%%%%%%%%%%%%%%%%%%%%
%%%%%%%%%%%%%%%%%%%%%%%%%%%%%%%%%%%%%%%%%%%%%%%%%%%%%%%%%%%%%%%%%

\begin{figure}[h]
    \centering
    \includegraphics[width=\textwidth]{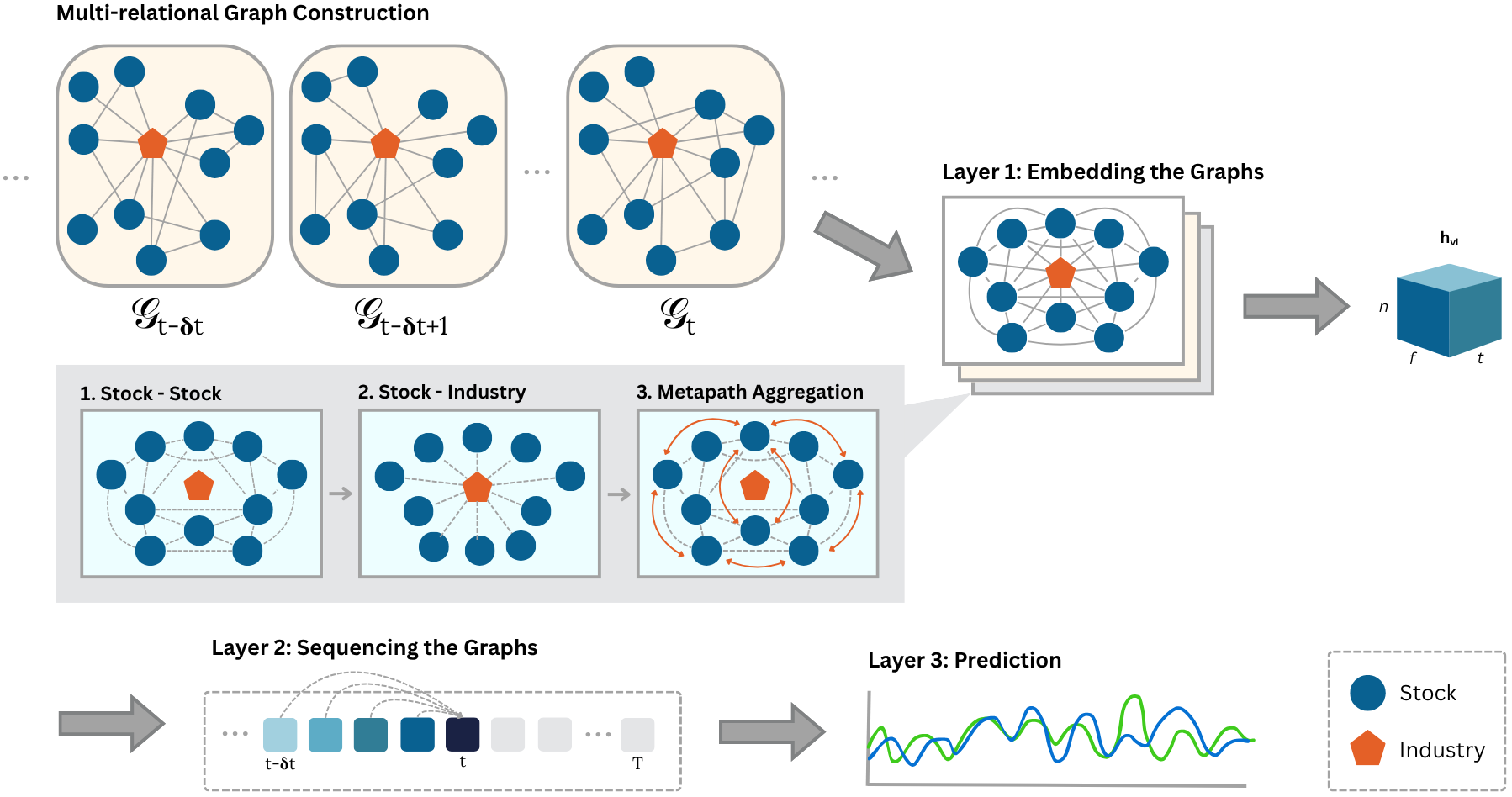}
    \caption{OmniGNN Model Overview}
    \label{fig:1}
\end{figure}

\subsection*{Layer 1: Embedding the Graphs}
\subsubsection*{Metapath Construction}
To encode the multi-relational nature of the financial market, we define a set of two metapaths $\mathcal{P} = \{\mathcal{SS}, \mathcal{SIS}\}$ that reflect distinct semantic relationships between stock nodes. Metapaths function as schema-level patterns that enable the model to aggregate messages not only from immediate neighbors but also from nodes that are semantically related through intermediate entities (e.g., shared industries). Each metapath $\mathcal{P}$ induces its own adjacency matrix $A_\mathcal{P}$ and edge attribute matrix $E_\mathcal{P}$. For the $\mathcal{SIS}$ metapath, we compute the adjacency as follows:

\begin{equation}
    A_{\mathcal{SIS}} = A_{\mathcal{SI}} \cdot A_{\mathcal{IS}} = A_{\mathcal{SI}} \cdot A_{\mathcal{SI}}^{\top}, \qquad A_{\mathcal{SIS}}\in\{0,1\}^{N\times N}.
\end{equation}

This operation connects stocks that share at least one industry, thereby modeling industry-based relational similarity. Edge features for $\mathcal{SIS}$ are similarly derived by averaging the edge features along the path.
% \begin{equation}
%     E_{\mathcal{SIS}}[i,j] = \frac{1}{2}\left(E_{\mathcal{SI}}[i,k] + E_{\mathcal{SI}}[j,k]\right),
% \end{equation}
% where $k$ indexes the shared industry node.

\subsubsection*{Graph Attention Mechanism}
    For a set of node features $\mathbf{h} = {\{\mathbf{h}_1, \mathbf{h}_2, \cdots, \mathbf{h}_N}\}$, $\mathbf{h}_i \in \mathbb{R}^F$, our goal is to obtain a context-aware node embedding that incorporates the most relevant information from neighboring nodes. To enable full pairwise attention, we adopt the architectural framework proposed by Graph Attention Networks \cite{3}. 

    In a single graph attention layer, we first apply a shared linear transformation, represented by the weight matrix $\mathbf{W}$, to both the node feature vectors and the edge attributes of the graph. This projects all inputs into a shared latent space, enabling the model to learn feature interactions in a comparable space. For each node pair $(v_i,v_j)$, the attention coefficient $\beta_{ij}$ is computed as follows: 
    \begin{equation}
    \beta_{ij} = a^{\top}(\mathbf{W} {\mathbf{h}_i} || \mathbf{W}{\mathbf{h}_j} || \mathbf{W} {\mathbf{e}_{ij}}),
    \end{equation} where $a$ is a learnable attention mechanism: $\mathbb{R}^{F'} \times \mathbb{R}^{F'} \rightarrow \mathbb{R}.$ The attention score captures the pairwise relevance of neighbor $v_j$'s node and edge features to $v_i$’s next state.
   
    After enabling nonlinearity through LeakyReLU, the attention coefficient is normalized across node features via softmax activation  $(\theta = 0.2)$:
     \begin{equation}
     \alpha_{ij} = \frac{\text{exp}(\text{LeakyReLU}(\beta_{ij}))}{\sum_{k \in \mathcal{N}_i} \text{exp}(\text{LeakyReLU}(\beta_{ij}))},
    \end{equation}
    where $\mathcal{N}_i$ represents the neighborhood of the node, determined by the adjacency matrix. 

    Finally, the resulting node embeddings $\mathbf{h'} = {\{\mathbf{h'}_1, \mathbf{h'}_2, \cdots, \mathbf{h'}_N}\}$, $\mathbf{h'}_i \in \mathbb{R}^{F'}$
    are updated by aggregating across $H$ attention heads via average pooling:
    \begin{equation} 
    \mathbf{h}_i' = \frac{1}{H}\sum_{k=1}^H\sum_{j \in \mathcal{N}_i} \alpha_{ij}^k \mathbf{W}^k \mathbf{h}_j \in \mathbb{R}^{H \cdot F}.
    \end{equation}

    \begin{figure}
        \centering
        \includegraphics[scale=0.35]{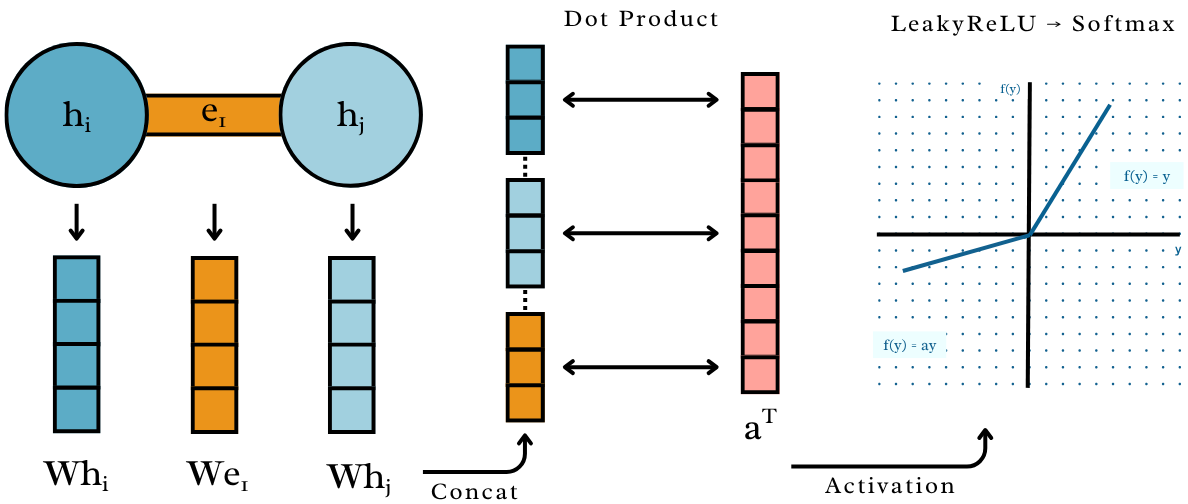}
        \label{fig:enter-label}
        \caption{Graph Attention Networks Diagram}
\end{figure}

    This step fuses multi-head context into the final embedding for each node, enriching it with structurally-aware information from its neighborhood.
    
This sequence is repeated for each time slice $[t-\delta_t,\cdots,t]$, producing a temporal sequence \hbox{$\mathbf{H}_{v,t-\delta_t:t} \in \mathbb{R}^{\delta_t\times d_h}$} for each node $v$, where $\delta_t$ is the window size.

%%%%%%%%%%%%%%%%%%%%%%%%%%%%%%%%%%%%%%%%%%%%%%%%%%%%%%%%%%%%%%%%%
%%%%%%%%%%%%%% LAYER 2: TEMPORAL LAYER %%%%%%%%%%%%%%%%%%%%%%%%%%
%%%%%%%%%%%%%%%%%%%%%%%%%%%%%%%%%%%%%%%%%%%%%%%%%%%%%%%%%%%%%%%%%

\subsection*{Layer 2: Sequencing the Graphs}
While the structural layer captures node representations that reflect their relationships within the corporate network on a given day $t$, these relationships are constantly evolving. Stock features like price, volatility, and news sentiment fluctuate daily, while institutional shareholders will enter or exit positions over time, reshaping the network. Therefore, it is essential to extract the temporal evolution of node and edge features from the graph snapshot sequence.

To accomplish this task, we employ the Transformer architecture with ALiBi to introduce an inductive bias toward recent graph snapshots \cite{11,12}. The Transformer has become a dominant architectural choice in many domains––particularly that of natural language processing and computer vision––due to its capacity for parallelization and contextual token embeddings. Unlike recurrent models such as GRUs and LSTMs, which process data sequentially by maintaining a hidden state variable, the Transformer uses self-attention to capture global context simultaneously across all time steps. This shift allows for more scalable and expressive modeling of temporal graph data. Its effectiveness for learning sequential data is repurposed here to model the temporal evolution of graph snapshots. 

The sequence of node embeddings over time $\mathbf{H}_{v,t-\delta_t:t}$ is passed through a Transformer with ALiBi for temporal modeling. The temporal attention computes the learned projections of embeddings into Query, Key, and Value spaces: 
\begin{equation}
    \mathbf{Q} = \mathbf{W}_\mathbf{Q}\mathbf{H}_{v,t-\delta_{t:t}}, \nonumber \qquad
    \mathbf{K} = \mathbf{W}_\mathbf{K}\mathbf{H}_{v,t-\delta_{t:t}}, \nonumber \qquad
    \mathbf{V} = \mathbf{W}_\mathbf{V}\mathbf{H}_{v,t-\delta_{t:t}},
\end{equation}
where $\mathbf{W_{\mathbf{Q}}}$, $\mathbf{W_{\mathbf{K}}}$, and $\mathbf{W_{\mathbf{V}}}$ are learned weight matrices. 
The temporal representation of the node is then obtained as follows: 
\begin{equation}
   \mathbf{Z} = \text{softmax}\left(\frac{\mathbf{QK}^{\top}}{\sqrt{d_k}} + m \cdot \mathbf{P} + \mathbf{M}\right)\mathbf{V},
\end{equation}
where $m\cdot \mathbf{P}$ is the ALiBi term with $m$ controlling the strength of the bias, and $\mathbf{M}$ acts as a causal mask, ensuring that the model attends only to past and present time steps.

\begin{figure}
        \centering
        \includegraphics[scale=0.3]{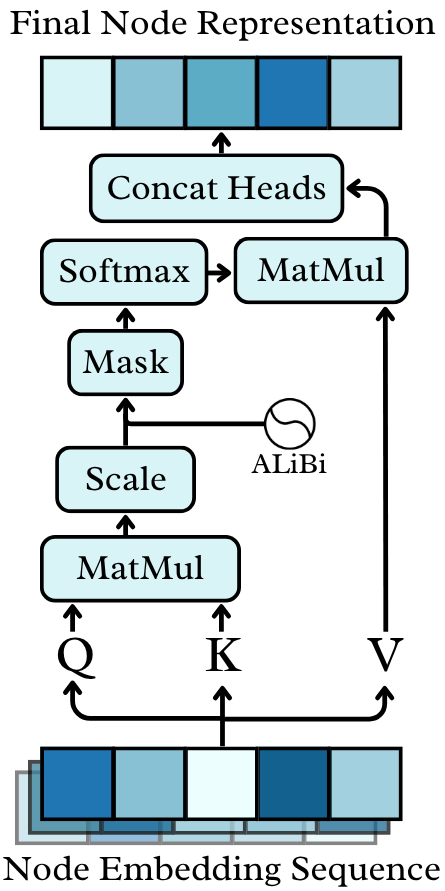}
        \label{fig:enter-label}
        \caption{Transformer Model Diagram}
\end{figure}

\subsection*{Layer 3: Prediction}
Each stock node is equipped with a dedicated linear prediction head that predicts a scalar \textbf{excess return} value. Concretely, we parameterize this using a distinct linear layer
\begin{equation}
    f_i(\cdot) = \mathbf{W}_i\mathbf{z}_{it}+\mathbf{b}_i
\end{equation}
for each stock node $i$, where $\mathbf{z}_{it}$ is the temporal embedding output produced by the Transformer encoder. 
% \begin{equation}
%     \hat{y}_{vt} = \mathbf{W}_1\mathbf{z}_{vt}+\mathbf{b}_1.
% \end{equation}
Ground truth is defined to be the excess return relative to the S\&P 500 index benchmark: 
\begin{equation}
    y_{it} = \frac{p_{i,t+1}\;-\;p_{i,t}}{p_{i,t}} - \frac{\texttt{SPX}_{t+1}\;-\;\texttt{SPX}_{t}}{\texttt{SPX}_{t}}.
\end{equation}

\section{Experimental Design}

\subsection*{Datasets}
We leverage the Bloomberg Terminal\footnote{Source: Bloomberg Finance L.P.} and London Stock Exchange\footnote{London Stock Exchange Group Data and Analytics.} to construct time-series datasets for 10 stocks in the Information Technology sector, encompassing their market performance, company valuation, financial news sentiment, ESG, and structural ownership features. The selection of 10 stocks is motivated by computational tractability; however, the framework is readily extensible to larger stock networks. Industry node features are derived from the market performance of the \texttt{XLK} Exchange-Traded Fund, which serves as a proxy for the Information Technology sector. Prior to analysis, all features are normalized, winsorized, and subjected to dimensionality reduction via Principal Component Analysis (PCA).

\subsection*{Backtesting}
We adopt a rolling window backtesting strategy to evaluate the model's performance across the varying market conditions represented in the period from January 4th, 2019 to December 15th, 2022. The full historical dataset is divided into overlapping windows that segment 6 months for training, 2 months for validation, and 2 months for testing. The Information Coefficient (IC), Information Ratio (IR), Cumulative Return (CR), and Precision@K (K=30\%) are computed for each cycle, and the window advances by the length of the testing period. 

\subsection*{Metrics}
We select the following standard financial metrics to assess the effectiveness of OmniGNN:
\begin{enumerate}
    \item \textbf{Information Coefficient (IC)} measures the \textit{Spearman rank correlation} between the model's predicted and ground truth excess returns;  
    \item \textbf{Information Ratio (IR)} measures the \textit{risk-adjusted performance} of the model, or the average return of the top-K predicted stocks each day normalized by its volatility;
    \item \textbf{Cumulative Return (CR)} is the \textit{accumulated return} over the test period achieved by taking daily long positions in the top-K\% of stocks ranked by predicted excess return;
    \item \textbf{Precision@K} measures the proportion of the top-K predicted stocks whose excess returns exceed the benchmark index.
\end{enumerate}

Model performance on the full historical dataset is evaluated by averaging the values of the model's performance metrics across all testing windows.

\section{Results}
\textbf{Baselines.} We benchmark OmniGNN against the GAT \cite{3} and Transformer \cite{12} models. The performance of our model, as well as select baseline models, are presented in Table 1 and illustrated in Figure 2. The values in parentheses denote the relative improvement of OmniGNN over each baseline model.

\textbf{Settings.} All models are built using PyTorch. The number of GNN layers is set to 3, and the hidden dimensions for each layer is x. We use H=x heads for the attention layer, while the Transformer uses x heads, x layers, x dropout and a window size of $\delta=$x. We use Adam as the optimizer, and set the hyperparameter weight decay to 1e-4 (l2 regularization), the learning rate to 1e-3, ($\beta$1, $\beta$2) to (0.9,0.999), $\epsilon$ to 1e-8. The batch size is set to 16. Each model instance is trained for 600 epochs with early stopping to ensure sufficient training and prevent overfitting. 
\begin{table}[h]
    \centering
    \begin{tabular}{lcccc}
        \hline
        Model & IC & IR & CR & Prec@K \\
        \hline
        Transformer & \makecell{0.0225 \\ (1.99)} & \makecell{0.0460 \\ (0.67)} & \makecell{0.0102 \\ (1.13)} & \makecell{0.5036 \\ (0.05)} \\
        \hline
        GAT         & \makecell{0.0209 \\ (2.22)} & \makecell{-0.0009 \\ (86.2)} & \makecell{-0.0028 \\ (8.79)} & \makecell{0.5041 \\ (0.04)} \\
        \hline
        \textbf{OmniGNN}     & \makecell{\textbf{0.0673} \\ }        & \makecell{\textbf{0.0767} \\ }        & \makecell{\textbf{0.0218} \\ }        & \makecell{\textbf{0.5266} \\ } \\
        \hline
    \end{tabular}
    \caption{Results of OmniGNN vs. Baseline model performance on the full historical dataset.}
    \label{tab:performance_comparison}
\end{table}

\begin{figure}[h]
    \centering
    \includegraphics[width=\textwidth]{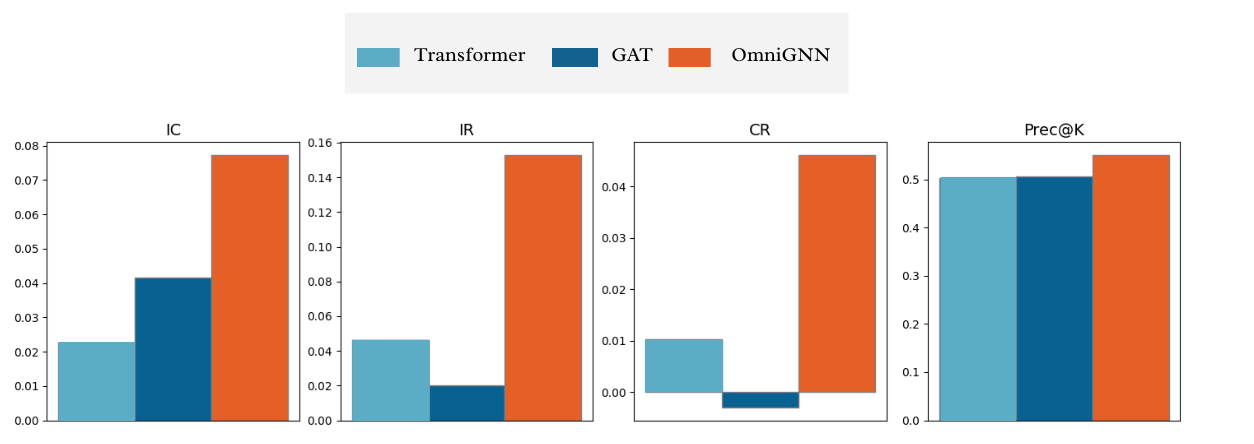}
    \caption{ Bar charts of OmniGNN vs. Baseline models}
    \label{fig:1}
\end{figure}
 The experiments demonstrate that OmniGNN outperforms the baseline models across the given performance assessment metrics. By incorporating both structural and temporal encoding, OmniGNN surpasses GAT by 2.22 and Transformer by 1.99, as measured by IC. Using the top-K portfolio strategy, OmniGNN achieves an average cumulative return of 0.0218 over the 2 month testing periods. The daily IC of 0.0673 indicates that the model generates a moderately meaningful predictive signal \cite{zhang2020information}. The robustness of the model is further demonstrated by its ability to generate accurate predictions for a basket of 10 technology stocks, which exhibit an average daily volatility of 0.0228—58\% higher than that of the \texttt{SPX} index over the same period. 
%Assuming that returns compound over time, the CR annualizes to roughly 13.8\%, compared to the 10.04\% CR of the \texttt{SPX} index over the same testing periods.
%OmniGNN's reported IR is 0.0767, which means that the model generates little positive risk-adjusted return.

\subsection*{Case Study: COVID-19}
To test the impact of the global node on model performance, we perform an ablation experiment by removing the industry node, the $\mathcal{E}_{\mathcal{SI}}$ edge, and the corresponding $\mathcal{SIS}$ metapath. Notably, the testing window covers the onset of the COVID-19 pandemic (March 4 – May 4, 2020), allowing us to evaluate model performance during a time of extreme market volatility.

\begin{table}[h]
    \centering
    \begin{tabular}{lcccc}
        \hline
        Metapaths & IC & IR & CR & Prec@K \\
        \hline
        $\mathcal{SS}$ & \makecell{-0.0745 \\ (0.38)} 
                       & \makecell{-0.0565 \\ (2.45)} 
                       & \makecell{-0.0275 \\ (2.23)} 
                       & \makecell{0.5050 \\ (0.11)} \\
        \hline
        $\mathcal{SS}+\mathcal{SIS}$ 
                       & \makecell{\textbf{-0.0465} \\ } 
                       & \makecell{\textbf{0.0819} \\ } 
                       & \makecell{\textbf{0.0338} \\ } 
                       & \makecell{\textbf{0.5590} \\ } \\
        \hline
    \end{tabular}
    \caption{Results of metapath ablation on key financial metrics during COVID-19. Parentheses indicate relative improvement of $\mathcal{SS}+\mathcal{SIS}$ over $\mathcal{SS}$.}
    \label{tab:metapath_ablation}
\end{table}

The inclusion of the industry node and its corresponding edges and metapaths leads to improvements across all metrics during the onset of the COVID-19 pandemic. This is likely because during the quarantine and ensuing financial crisis that characterized the pandemic, macroeconomic shocks impacted companies through their industry-specific risk exposures, captured by the $\mathcal{SIS}$ metapath. 

\section{Future Work}
Building on the insights from our current model, there remain several promising directions we aim to explore in future work. These include architectural modifications, alternative supervision signals, and improved graph construction strategies that may further enhance predictive performance and interpretability.
\begin{itemize}
    \item \textbf{Time-Lag Incorporation.} In predicting future stock price behavior, OmniGNN relies on technical analysis; however, this approach doesn't account for the time lag inherent to some financial data, like news sentiment scores. For example, investor sentiment expressed in the news on a given day will not necessarily affect stock prices immediately. Thus, incorporating lag-adjusted features may improve the predictive accuracy of OmniGNN \cite{17}. 
    \item \textbf{Edge Attributes.} Sourcing broader relationship types (industry, subsidiary, people, and product relations) through open-source knowledge graphs can be used to increase the dimensions of edge attributes. This approach, as explored by IBM Research Japan, enhances the expressiveness of the graph structure \cite{6}. Subsequently, dimensionality reduction techniques such as PCA can be applied to distill more meaningful and compact representations of inter-entity relationships.
    \item \textbf{Advanced Sentiment Analysis.} OmniGNN utilizes the text processing library \textbf{TextBlob} \footnote{https://textblob.readthedocs.io/en/dev/} to obtain polarity scores for each news article. However, the default models in \textbf{TextBlob} are not purpose-built for financial news, so the use of more specialized natural language processing techniques may yield less noisy, more informative features. 
    \item \textbf{New Node Types.} The next steps also include designing new node types, such as nodes that store several macroeconomic indicators or determine explicit market regimes.
\end{itemize}

\section{Conclusion}
Our proposed model, OmniGNN, makes several contributions to the GNN literature. With multi-dimensional metapaths defining longer-hop relationships through the heterogeneous network schema, OmniGNN applies attention weights to learn structurally meaningful representations of stock nodes while filtering out noise from less informative connections. As shown through ablation experiments, these new paths improve the predictive performance of downstream tasks like node-level regression. OmniGNN repurposes the innovative Transformer mechanism to sequence graph snapshots, which transforms the hidden embeddings into ones that account for longer-range temporal dependencies between trading days. Finally, OmniGNN's structural innovation, the global node intermediary, densifies graphs with a star topology overlay, enabling efficient message propagation during periods of global disruption. Our results show that OmniGNN achieves consistent predictive power for the daily excess returns for 10 relatively volatile tech stocks during a 4-year period with greatly varied market regimes.

OmniGNN's architecture also allows for generalization to larger universes of stocks. Future work will focus on extending our experiments to incorporate entire corporate networks, such as the entire Information Technology sector or S\&P 500.

\section*{Acknowledgments}
This research was conducted as part of the Columbia
Summer Undergraduate Research Experiences in Mathematical Modeling program, hosted and supported by the Columbia University Department of Mathematics. We are grateful to Professor George Dragomir, Professor Dobrin Marchev, and Vihan Pandey for their support throughout the project. 

%%%%% BIBLIOGRAPHY %%%%%%
%% \printbibliography
\printbibliography

\end{document}